\documentclass[fleqn,10pt]{wlscirep}
\usepackage{lineno}
\usepackage[utf8]{inputenc}
\usepackage[T1]{fontenc}
\usepackage{url}
\usepackage[capitalise, noabbrev]{cleveref}
\usepackage{multirow}
\usepackage{booktabs}
\usepackage{amsmath}
\usepackage{graphicx}
\usepackage{enumitem}
\usepackage{subcaption}

\newcommand{\name}{\emph{Latent Boost}}

\title{Enhancing Interpretability Through Loss-Defined Classification Objective in Structured Latent Spaces}
\author[1]{Daniel Geißler}
\author[1,2,*]{Bo Zhou}
\author[1]{Mengxi Liu}
\author[1,2]{Paul Lukowicz}

\affil[1]{German Research Center for Artificial Intelligence (DFKI), Kaiserslautern, Germany}
\affil[2]{University of Kaiserslautern-Landau (RPTU), Kaiserslautern, Germany}
\affil[*]{bo.zhou@rptu.de}

\begin{abstract}
Supervised machine learning often operates on the data-driven paradigm, wherein internal model parameters are autonomously optimized to converge predicted outputs with the ground truth, devoid of explicitly programming rules or a priori assumptions.
Although data-driven methods have yielded notable successes across various benchmark datasets, they inherently treat models as opaque entities, thereby limiting their interpretability and yielding a lack of explanatory insights into their decision-making processes.
In this work, we introduce \name{}, a novel approach that integrates advanced distance metric learning into supervised classification tasks, enhancing both interpretability and training efficiency.
Thus during training, the model is not only optimized for classification metrics of the discrete data points but also adheres to the rule that the collective representation zones of each class should be sharply clustered.
By leveraging the rich structural insights of intermediate model layer latent representations, \name{} improves classification interpretability, as demonstrated by higher Silhouette scores, while accelerating training convergence.
These performance and latent structural benefits are achieved with minimum additional cost, making it broadly applicable across various datasets without requiring data-specific adjustments.
Furthermore, \name{} introduces a new paradigm for aligning classification performance with improved model transparency to address the challenges of black-box models.

\end{abstract}

\begin{document}

\flushbottom
\maketitle
%
%
\thispagestyle{empty}


\section{Introduction}
Machine learning models are increasingly used across a wide range of domains, but their reliance on black-box architectures often obscures the internal decision-making processes, raising concerns about transparency and reliability \cite{qamar2023understanding,geissler2023latent}.
Since Deep Neural Networks increasingly expand into sensitive domains such as medical, autonomous driving, or the avionics sector, the inability to interpret model decisions creates significant barriers to deployment and certification \cite{bello2024towards}. 
Trustworthy machine learning requires models not only to perform well but also to offer explanations that align with human understanding, ensuring that decisions are robust and free of unintended biases \cite{Eshete2021Making}.
This limitation is particularly evident in traditional data-driven classification training, which focuses solely on optimizing classification scores for discrete datasets while neglecting the structural organization of clusters within the continuous latent representation \cite{csahin2024unlocking}.

As a promising solution, latent representations, extracted from intermediate outputs of neural networks, contain rich information about the propagated data \cite{crabbe2021explaining}. 
However, without explicit guidance during training, these representations often lack cohesion, with semantically similar instances failing to cluster effectively. 
This lack of organization can undermine the interpretability of latent spaces, as well as the ability of downstream tasks to utilize the encoded information efficiently \cite{Esser2020A}. 
Models trained in this way may struggle to generalize beyond the training data, especially in the presence of domain shifts through out-of-distribution deployments \cite{lu2021invariant}. 
Addressing these deficiencies is critical for building robust systems capable of handling the ubiquitous challenges of real-world scenarios \cite{Rybakov2020Learning}.
This lack of focus on latent structure not only limits interpretability but can also impede the model’s ability to generalize effectively across varying domains and datasets \cite{kotyan2024linking}.
To address these challenges, we propose an innovative approach that explicitly integrates the classification objective into the latent representation through distance metric learning.

\begin{figure}[ht] 
    \centering
    \includegraphics[width=0.9\textwidth]{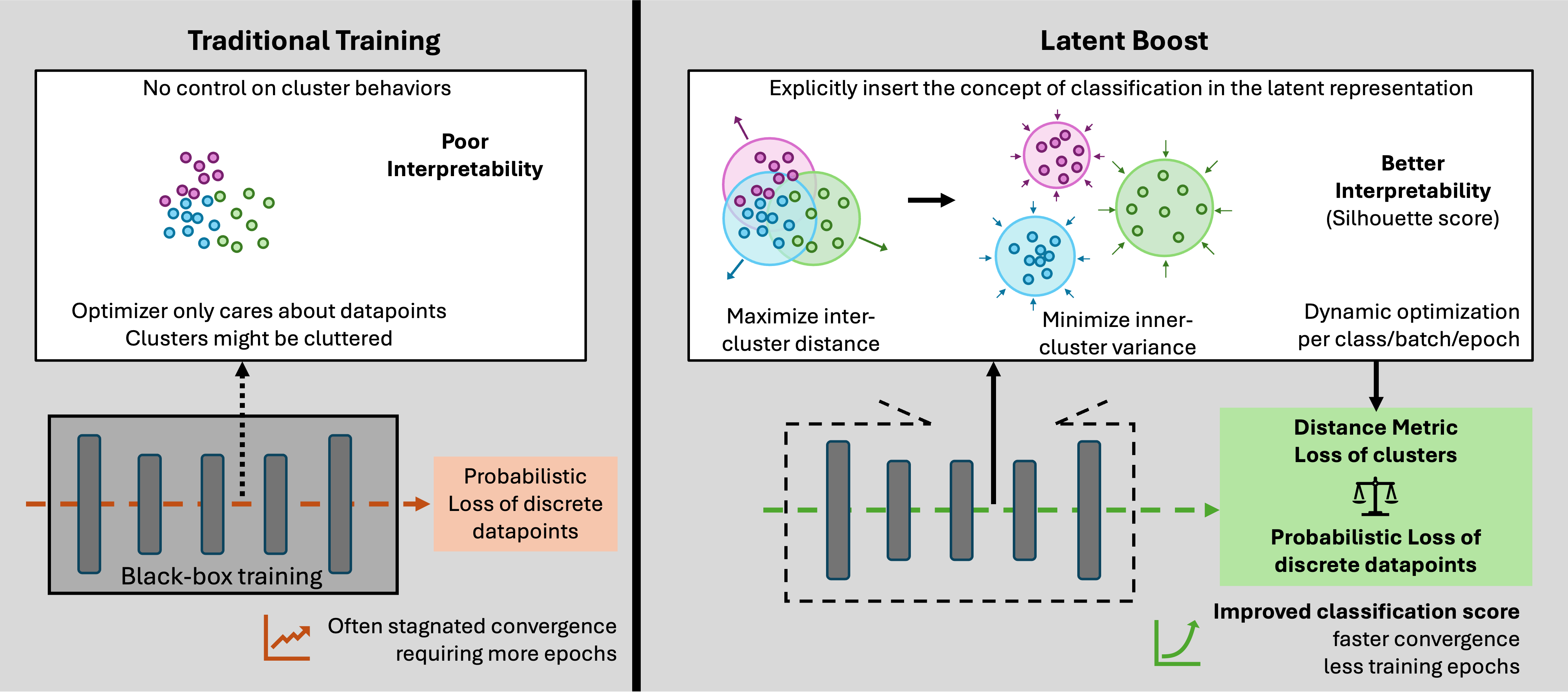} 
    \caption{Oppose to traditional training, relying on probabilistic loss only, \name{} injects distance metric information, obtained from the model's hidden latent representations, as addition into the training through balanced weighted sum equations.}
    \label{fig:teaser}
    \vspace{-1em}
\end{figure}

In this work, we introduce a novel method, \name{}, that seamlessly combines latent cluster distance metrics with probabilistic training as motivated in \Cref{fig:teaser}, fundamentally transforming the paradigm of structured latent representations to improve both transparency and performance.
While conventional probabilistic approaches typically center on individual data samples, they often overlook the intricate relationships among data points, particularly within clusters, where the interdependencies and structural nuances are essential for capturing the underlying distribution patterns \cite{wu2019towards}.
In other words, while the end-to-end classification performance might be satisfactory, the results like the F1-Score are derived from discrete independent data points. 
Internally, the collective data may not form clear clusters, or the formed clusters of different classes could still be cluttered in the continuous latent representation, as the training processes only optimize for the probabilistic loss of the discrete dataset.
In contrast, our innovative approach guarantees that semantically similar data points are thoroughly aligned in proximity, while dissimilar points are distinctly separated, thereby enhancing the model’s capacity for nuanced differentiation.
This not only improves the interpretability of the learned latent representations but also provides better alignment with the underlying data distribution, ensuring more meaningful and reliable predictions.

Although distance metric learning has established its efficacy in Machine Learning—particularly in clustering and pre-training tasks \cite{kulis2013metric,kaya2019deep}—its application in classification has been limited to simpler models like K-nearest Neighbors \cite{cover1967nearest} and Support Vector Machines \cite{cortes1995support}. 
We propound that intermediate latent representations hold critical structural information essential for class distributions, and enhancing their cluster separation can significantly boost classification performance. To the best of our knowledge, \name{} is the first method to integrate distance metrics into the classification loss function, empowering the model to cultivate more meaningful and discriminative features without requiring explicit supervision.
By incorporating distance-based loss into the weighted training process, \name{} bridges the gap between structural clustering methods and probabilistic classification, delivering a holistic framework for supervised learning.

Our key contributions include:
\begin{itemize}
    \item We incorporate distance metric learning, traditionally developed for unsupervised clustering, into supervised classification through a weighted sum loss equation. This allows the model to be optimized for both classification metrics and interpretable latent representations during the standard training process.
    \item  We propose a novel distance-based loss \name{} specifically for supervised classification, addressing previously overlooked nuances with dynamic adaptation and discriminative information density.
    \item We demonstrate \name{} is a simple yet efficient approach to enhance performance and interpretability while shrinking computational demand through faster convergence.
\end{itemize}

\section{Related Work}

Distance Metric Learning has emerged as a crucial area in Machine Learning, offering a wide range of techniques aimed at improving performance in tasks of primarily unsupervised clustering and retrieval of latent representation information in general \cite{kulis2013metric, wang2015survey}.
Such loss functions generally focus on minimizing intra-cluster variance and maximizing inter-cluster distances by learning a latent representation where similar points are closer together, and dissimilar points are further apart \cite{kaya2019deep}. 
These functions penalize the model until the latent representation aligns with the desired distance metric priorities. 

The structure and information density in latent representations also improve interpretability.
Discriminative Dimension Selection can be utilized to enhance the interpretability and clustering performance, especially for K-means clustering by selectively retaining relevant features \cite{lian2024discriminative}. 
Similarly, works on adaptive feature selection and optimization have focused on developing efficient strategies to reduce the burden of complex dimensionality, improving clustering accuracy across several benchmarks \cite{zhou2024multi,geissler2024power}.
Oppose to utilizing the Euclidean distance information, a boosting algorithm to effectively learn Mahalanobis distance metrics was proposed by Chang et al., demonstrating its effectiveness on popular datasets to capture intrinsic distance relationships \cite{Chang2012A}.
Mahalanobis-based techniques have been widely adopted, with approaches such as Large Margin Nearest Neighbor \cite{weinberger2009distance} maximizing the margin between different classes, and Information-Theoretic Metric Learning \cite{davis2007information}, which minimizes the relative entropy between distance distributions.

Pairwise and Contrastive loss functions have shown remarkable improvements in distance metric learning, such as the Contrastive loss function for learning dimensionality-reducing embeddings \cite{hadsell2006dimensionality}. 
This principle has been adopted in domains like zero-shot learning \cite{wang2017zero}, cross-modal retrieval \cite{wang2017adversarial}, and large-scale face recognition \cite{liu2017sphereface}. 
Contrastive learning has also been used in self-supervised settings, where methods such as SimCLR \cite{chen2020simple} leverage this loss to learn robust features without labels.
The promising approaches of triplet-based methods for distance metric learning have been challenging to optimize due to the need for finding informative triplet anchor points \cite{Do2019A}. 
To address this, semi-hard triplet mining methods have been developed, leading to more efficient training \cite{schroff2015facenet}. 
Advanced sampling strategies have been proposed to improve the performance of triplet-based learning systems \cite{hermans2017defense}. 
Additionally, using proxy points to approximate original data points, as shown by  Movshovitz-Attias et al., further improves convergence and stability in Triplet loss optimization \cite{Movshovitz-Attias2017No}.
Magnet loss has gained attention in recent years as an effective method for distance metric learning, particularly in dealing with high-dimensional data where traditional losses like Triplet loss face challenges \cite{rippel2015metric}.
Yu et al. introduce Semantic Drift Compensation as part of Magnet loss as a method to address catastrophic forgetting in class-incremental learning by estimating and compensating for the feature drift of previous tasks, significantly improving performance in embedding networks without requiring exemplars \cite{yu2020semantic}.
By focusing on the distribution of latent representations within each class, Magnet loss works to minimize the overlap between class clusters, thus providing better discrimination, especially in face identification in noisy and large-scale datasets \cite{deng2020sub}.

Metric learning with constraints has led to significant advances, with pairwise constraints being integrated into methods like Constrained Clustering via Metric Learning \cite{bilenko2004integrating}.
Similarly, Ding et al. extended this concept into semi-supervised clustering, demonstrating the utility of leveraging partial label information \cite{ding2007adaptive}. 
In sparse and high-dimensional data scenarios, sparse subspace clustering has been a successful approach \cite{liu2010robust}.

Fairness in distance metric learning has also become a crucial area of focus, with works exploring how adversarial techniques can be used to ensure fairness in learned distance metrics \cite{lahoti2020fairness}. 
These methods prevent demographic bias and ensure equitable performance across different groups, a significant consideration for real-world applications.

Methods like DeepCluster \cite{caron2018deep} and cluster-based contrastive learning \cite{caron2020unsupervised} demonstrate how metric learning can generate meaningful representations for downstream tasks such as image retrieval. 
The innovation in these approaches lies in integrating clustering with deep learning techniques to build robust data representations. 
Hybrid models that integrate different learning techniques have also gained attention. 
For example, Lee et al. \cite{lee2018stacked} employed stacked attention networks for cross-modal tasks, combining textual and visual data within joint latent spaces to enable multi-modal learning.
However, those methods rest on unsupervised clustering techniques, wherefore they are unsuitable for our supervised multi-class classification approach.

\section{Preliminaries}

\subsection{Distance Metric Learning Theoretical Background}
Based on the previously referenced works, we identified the Contrastive, Triplet, N-pair, and Magnet loss as the currently most relevant and recent choices wherefore we outline their concepts in the following:

\textbf{Contrastive Loss} is one of the simplest and most widely used loss functions for distance metric learning, introduced in the context of training Siamese networks \cite{chopra2005learning}. 
The goal of Contrastive loss is to minimize the distance between pairs of samples that are similar and maximize the distance between pairs of samples that are dissimilar up to a certain margin.
The Contrastive loss function is formulated in \Cref{eq:contrast}, where \( y_i \in \{0, 1\} \) is a binary indicator of similarity, with \( y_i = 1 \) for similar pairs and \( y_i = 0 \) for dissimilar ones. 
The Euclidean distance \( d_i \) is calculated between the latent representations of the two samples.
The equation is applied across the total number of pairs \( N \).
Additionally, \( m \) works as a penalization, defining the margin as the minimum distance for dissimilar pairs.
\begin{equation}
\label{eq:contrast}
\mathcal{L}_{\text{contrast}} = \frac{1}{2N} \sum_{i=1}^{N} \left( y_i \cdot d_i^2 + (1 - y_i) \cdot \max(0, m - d_i)^2 \right)
\end{equation}
\textbf{Triplet Loss} was used by the FaceNet model architecture, introducing the triplets of samples \cite{schroff2015facenet} as shown in \Cref{eq:triplet}.
An anchor \( a_i \), a positive sample \( p_i \) , and a negative sample \( n_i \) form one triplet.
The goal is to ensure that the distance between the anchor and the positive sample is smaller than the distance between the anchor and the negative sample by at least a margin \( \alpha \).
Compared to the Contrastive loss, the Triplet loss is based on the anchor and is not calculated solely on the pairwise samples.
The Euclidean distance is calculated between the positive and negative sample to its anchor with \( m \) as the margin term.
\begin{equation}
\label{eq:triplet}
\mathcal{L}_{\text{triplet}} = \frac{1}{N} \sum_{i=1}^{N} \max \left( 0, d(a_i, p_i) - d(a_i, n_i) + m \right)
\end{equation}

\textbf{N-pair Loss} shares similarity with Triplet loss and aims to converge more stable across the clusters \cite{sohn2016improved}.
Instead of using a single negative sample per triplet, N-pair loss optimizes the distance between the anchor and the positive sample while contrasting it with multiple negatives simultaneously.
N-pair loss is particularly effective in multi-class classification tasks for large-scale datasets.
The N-pair loss is defined in \Cref{eq:npair} (\( \mathbf{f}_i \) is the embedding of the anchor sample, \( \mathbf{f}_i^+ \) and \( \mathbf{f}_i^-\) are positive and negative samples)
\begin{equation}
\label{eq:npair}
\mathcal{L}_{\text{N-pair}} = \frac{1}{N} \sum_{i=1}^{N} \log \left( 1 + \sum_{i^+ \neq i^-} e^{ \mathbf{f}_i^T \mathbf{f}_i^- - \mathbf{f}_i^T \mathbf{f}_i^+ } \right)
\end{equation}

\textbf{Magnet Loss}, introduced by Rippel et al. \cite{rippel2015metric}, is designed to address the challenges of high-dimensional and complex data distributions by grouping data into clusters instead of focusing on pairwise or triplet calculations.
To benefit from cluster-based information, Magnet loss penalizes a sample based on its distance to the centroid of the correct cluster and the other false clusters.
The behavior of a Magnet is obtained by accounting for both intra-class compactness and inter-class separation.
The Magnet loss function is formulated in \Cref{eq:magnet_base}, where \( r_n \) is the latent representation of the sample, \( \mu(r_n) \) is the mean of the cluster containing \( r_n \), and \( \mu_c \) are the means of other clusters. 
\( \sigma \) is the standard deviation, \( \alpha \) is a margin term. 
\( N \) is the total number of samples whereas \( K \) represents the number of clusters.
\begin{equation}
\label{eq:magnet_base}
\mathcal{L}_{\text{magnet}} = \frac{1}{N} \sum_{n=1}^{N} \left\{ - \log \left( \frac{e^{-\frac{1}{2\sigma^2} \|\mathbf{r}_n - \boldsymbol{\mu}(r_n)\|^2_2 - \alpha}}{\sum_{c \neq C(r_n)} \sum_{k=1}^K e^{-\frac{1}{2\sigma^2}  \|\mathbf{r}_n - \boldsymbol{\mu}^c_k\|^2_2}} \right) \right\}
\end{equation}

\subsection{Incorporating Distance Metrics in Multi-Class Classification}

As the first novel contribution of this work, we incorporate distance metric information into supervised classification with probabilistic loss through a weighted sum loss function.
Such distance metrics have been hitherto primarily used for unsupervised clustering.
\Cref{eq:loss_function1} outlines the total loss calculation on which our experiments rest.
We introduce the hyperparameter $\lambda$ to balance the weight between the two loss components.
The range of $\lambda$ can be continuously selected between 0 to 1, with 0 giving full weight to the probabilistic loss whereas 1 gives full weight to the distance metric loss.
In order to isolate the influence of $\lambda$ and the effect of each selected distance metric, we fixed the probabilistic loss within each experiment across epochs to shrink the overall experiment complexity.
Moreover, with this approach the loss functions balance stays constant across the experiment, preventing any training inconsistencies through unexpected loss manipulations.
Additionally, we selected the well-established soft-max cross-entropy as a counterpart, which matches the idea of utilizing the weighted sum equation for multi-class classification.
\begin{equation}
\label{eq:loss_function1}
\begin{gathered}
    \mathcal{L}_{total} = \lambda \cdot \mathcal{L}_{\text{dist}} + (1 - \lambda) \cdot \mathcal{L}_{\text{cross-entropy}} \\
    \text{with } \mathcal{L}_{\text{dist}} \in \{ \mathcal{L}_{\text{contrast}}, \mathcal{L}_{\text{triplet}}, \mathcal{L}_{\text{N-pair}}, \mathcal{L}_{\text{magnet}}\}, \quad
    \mathcal{L}_{\text{cross-entropy}} = -\sum_{\forall x} p(x) \log(q(x))
\end{gathered}
\end{equation}

\subsection{Experiment Setup}
\label{sec:exp_setup}

We selected three different experiment setups of varying complexity to explore the effects of the previously introduced distance metric losses through the weighted sum equations.
The datasets used include Fashion MNIST, a collection of grayscale images representing 10 fashion categories, chosen for its simplicity and suitability for validating foundational improvements; CIFAR-10, a dataset of natural color images spanning 10 object classes, selected to assess performance on more complex visual data; and CIFAR-100, which significantly increases the challenge with its 100 fine-grained categories of color images, providing a rigorous evaluation of scalability and the ability to handle diverse, high-dimensional latent spaces.
The model architecture trained on Fashion MNIST \cite{xiao2017fashion} is a convolutional neural network (CNN) adapted for grayscale images.
The network starts with a convolutional layer that takes single-channel (grayscale) 28x28 images and outputs a set of feature maps, followed by ReLU activation and max-pooling. 
The model employs two convolutional layers, with each followed by pooling and dropout of 25\% to mitigate overfitting. 
The resulting feature maps are flattened and passed through fully connected layers.
From the flattening layer, we extract the latent features.
The final fully connected layer outputs the class probabilities for the 10 fashion categories.

The model architecture used for CIFAR-10 \cite{krizhevsky2009learning} is based on the VGG-16 network \cite{simonyan2014very}. 
VGG-16 is a deep convolutional network known for its success in image classification tasks on complex color images like CIFAR-10. 
This architecture consists of 16 layers, with 13 convolutional layers interspersed with ReLU activations and max pooling to downsample the feature maps, followed by three fully connected layers. 
The network extracts progressively richer features from the images, which are then classified into one of the 10 classes.

The model trained on CIFAR-100 \cite{krizhevsky2009learning} is based on the ResNet-50 architecture \cite{he2016deep} and comprises several key layers to efficiently process the input images. 
The architecture begins with a modified convolutional layer that accepts three input channels and employs a kernel size of 3 times 3, followed by batch normalization and ReLU activation. 
It includes four residual blocks (layer1 to layer4), each containing a series of convolutional layers, batch normalization, and skip connections to enhance gradient flow. The network concludes with an average pooling layer, which is flattened to extract the latent representations, and a fully connected layer that outputs the final class predictions, specifically tailored for the CIFAR-100 classification task.
For each experiment, we extracted the latent representation after the convolutional part in the network architecture, meaning just before the classification section.
For the CNN the latent representation has a dimension of 64, the VGG-16 model obtains 512 dimensions on that layer and the ResNet-50 generates a latent representation of 2048 dimensions.
The choice to extract latent features from layers just before the classification head is based on their ability to capture high-level, task-specific representations that balance discriminative power and compactness, as these layers contain semantically rich information essential for classification tasks. 
Earlier layers typically emphasize low-level features and may lack the necessary information density required for effective downstream processing, making deeper layers more suitable.

Each setup was implemented in PyTorch \cite{NEURIPS2019_9015} with a learning rate scheduler that reduced the learning rate by factor 5 when a plateau was reached after 10 epochs.
The Adam optimizer was utilized for each experiment to efficiently work out the gradients \cite{kingma2014adam}.
We applied early stopping metrics with 20 epochs patience based on the validation accuracy to additionally compare the convergence speed next to the classification performance.
Finally, we trained each experiment setup five times to provide meaningful results.
We assigned each trial a different random seed, which we kept constant within the trials different runs across the variation of lambda to balance the distance metric with the classic cross-entropy.

\subsection{Superiority of Magnet Loss}
\label{prelim_evaluation}
For our preliminary evaluation, we utilize the original hyperparameter settings for each distance metric loss function as shown in \Cref{tab:hyperparameters} and balance them in the weighted sum loss.
For Contrastive loss, the positive and negative margins are set to 0.0 and 1.0, respectively, which determine the threshold for distinguishing between positive and negative pairs. 
The Triplet loss employs a margin of 0.05 and considers all possible triplets per anchor, ensuring a comprehensive evaluation of relative distances in the latent representations. 
N-Pair loss utilizes a MeanReducer as its reducer, which averages the distances, while Magnet loss is configured with an $\alpha$ value of 1.0, controlling the aggressiveness for forcing the latent space separation. 
These initial values were chosen based on established practices in the literature to ensure a fair comparison independently of the distance metric initialization or adaptation.

\begin{table}[ht]
\footnotesize
    \centering
    \begin{tabular}{c|c|c}
    \hline
        \textbf{Loss Function} & \textbf{Hyperparameter} & \textbf{Initial Value} \\
        \hline
        \multirow{2}{*}{Contrast} & positive margin & 0.0 \\
        & negative margin & 1.0 \\
        \hline
        \multirow{2}{*}{Triplet} & margin & 0.05 \\
        & triplets per anchor & all \\
        \hline
        N-Pair& reducer & MeanReducer \\
        \hline
        Magnet& $\alpha$ & 1.0 \\
        \hline
    \end{tabular}
    \caption{Original hyperparameter setting for each distance metric utilized in the preliminary experiment.}
    \label{tab:hyperparameters}
    \vspace{-1em}
\end{table}

The results of varying the hyperparameter $\lambda$ across the different loss functions are reported across our three experiment setups, comparing the accuracy and Micro-F1 Scores.
We conducted each experiment across the range of 0.1 to 0.9 while keeping the value constant across each run.
In \Cref{tab:baseline}, we present the most relevant $\lambda$ values to compare and recognize trends.
The aim of this evaluation is to highlight the overall potential of outperforming the baseline performance of the model through latent distance information.
The results show that adjusting $\lambda$ can improve performance, especially when applying the Magnet loss. 
On Fashion MNIST, Magnet loss outperforms other methods, achieving 89.52\% accuracy and 0.8946 Micro-F1 scores at $\lambda$ = 0.75.
On CIFAR-10, Magnet loss also achieves the highest results, with an accuracy of 87.36\% and a Micro-F1 score of 0.8738 at $\lambda$ = 0.75.
However, on CIFAR-100, the results were less conclusive due to its high complexity, but Magnet loss still showed some improvement over the baseline. 
Overall, the Magnet loss is the most robust existing method across our experiments, especially when selecting $\lambda$ values in the range 0.5 to 0.9, since it properly improves classification performance across datasets by leveraging latent information and reducing intra-class variability with proper priority.

\begin{table}[!ht]
\centering
\footnotesize

\begin{tabular}{c|c|cc|cc|cc}
\hline
\multirow{2}{*}{\textbf{Loss Type}} & \multirow{2}{*}{\textbf{$\boldsymbol{\lambda}$}} & \multicolumn{2}{c|}{\textbf{Fashion MNIST}} & \multicolumn{2}{c|}{\textbf{CIFAR-10}} & \multicolumn{2}{c}{\textbf{CIFAR-100}} \\
\cline{3-8}
& & \textbf{Accuracy} & \textbf{Micro-F1} & \textbf{Accuracy} & \textbf{Micro-F1} & \textbf{Accuracy} & \textbf{Micro-F1} \\
\hline
Baseline & 0.0  & \textbf{88.59 ± 0.15} & \textbf{0.8867 ± 0.0020} & \textbf{85.88 ± 0.40} & \textbf{0.8586 ± 0.0042} & \textbf{61.77 ± 0.49} & \textbf{0.6163 ± 0.0064} \\
\hline
\multirow{3}{*}{Contrast} 
    & 0.25 & 88.83 ± 0.08 & 0.8884 ± 0.0015 & 86.01 ± 0.99 & 0.8600 ± 0.0102 & 61.04 ± 0.81 & 0.6088 ± 0.0087 \\
    & 0.5  & 88.57 ± 0.14 & 0.8850 ± 0.0019 & 86.87 ± 0.56 & 0.8693 ± 0.0049 & 60.00 ± 0.71 & 0.5984 ± 0.0087 \\
    & 0.75 & 87.79 ± 0.25 & 0.8773 ± 0.0022 & 84.74 ± 0.44 & 0.8479 ± 0.0045 & 54.91 ± 0.71 & 0.5482 ± 0.0071 \\
\hline
\multirow{3}{*}{Triplet} 
    & 0.25 & 89.13 ± 0.18 & 0.8909 ± 0.0011 & 86.81 ± 0.25 & 0.8667 ± 0.0025 & 62.36 ± 0.91 & 0.6213 ± 0.0070 \\
    & 0.5  & 89.13 ± 0.34 & 0.8915 ± 0.0036 & 86.88 ± 0.64 & 0.8683 ± 0.0059 & 60.41 ± 0.51 & 0.6022 ± 0.0041 \\
    & 0.75 & 88.97 ± 0.33 & 0.8900 ± 0.0038 & 86.95 ± 0.42 & 0.8690 ± 0.0039 & 56.13 ± 0.76 & 0.5597 ± 0.0082 \\
\hline
\multirow{3}{*}{N-pair}  
    & 0.25 & 88.85 ± 0.07 & 0.8883 ± 0.0010 & 84.52 ± 0.71 & 0.8446 ± 0.0067 & 61.54 ± 0.83 & 0.6139 ± 0.0075 \\
    & 0.5  & 89.25 ± 0.09 & 0.8923 ± 0.0005 & 86.43 ± 0.78 & 0.8643 ± 0.0081 & 60.13 ± 0.94 & 0.5991 ± 0.0075 \\
    & 0.75 & 88.71 ± 0.53 & 0.8869 ± 0.0044 & 85.91 ± 0.40 & 0.8583 ± 0.0042 & 47.73 ± 1.79 & 0.4694 ± 0.0168 \\
\hline
\multirow{3}{*}{Magnet}  
    & 0.25 & 89.27 ± 0.29 & 0.8928 ± 0.0023 & 86.91 ± 0.69 & 0.8692 ± 0.0073 & 62.43 ± 0.38 & 0.6242 ± 0.0039 \\
    & 0.5  & 89.07 ± 0.21 & 0.8911 ± 0.0022 & 86.72 ± 1.77 & 0.8671 ± 0.0174 & \textbf{62.75 ± 0.51} & \textbf{0.6256 ± 0.0057} \\
    & 0.75 & \textbf{89.52 ± 0.34} & \textbf{0.8946 ± 0.0040} & \textbf{87.36 ± 0.82} & \textbf{0.8738 ± 0.0081} & 62.66 ± 0.54 & 0.6242 ± 0.0045 \\
\hline
\end{tabular}
\caption{Baseline comparison of accuracy ($\uparrow$) and Micro-F1 score ($\uparrow$) across three datasets using different loss functions and $\lambda$ values. The baseline ($\lambda$=0.0) represents the standard model without additional loss integration.}
\label{tab:baseline}
\vspace{-1em}
\end{table}

\section{Latent Boost Evolution}

\label{sec:latentboost}
Among the hitherto loss functions, Magnet loss provides superior potential in enhancing classification performance, wherefore we selected the Magnet loss as a basis for our \name{} approach to incorporate distance with classification metrics.
However, those distance metric losses were initially developed for unsupervised clustering.
Although we have repurposed them for supervised classification through \cref{eq:loss_function1}, there is room for improvement, as for the classification objective, they have several limitations and overlooks certain nuances.
Specifically, no modifications or optimizations of the hyperparameters associated with the Magnet loss have been explored so far. 
Additionally, in the original formulation, the distance metric loss is computed across the full latent space. 
While this approach captures the complete available information, this might introduce noise and excessive computational demand if the latent representation includes dimensions that do not yet provide beneficial information.
The following sections introduce our evolutionary development of \name{}, targeting the maximization of efficiency and explainability.

\subsection{Condensed Distance Information}
As a first step, we propose incorporating a dimensionality reduction step before calculating the \name{} batch loss by applying Principal Component Analysis (PCA) to reduce the dimensions of the latent vectors. 
PCA was selected for its balance of simplicity, computational efficiency, and ability to retain the maximum variance in the data, making it particularly suitable for preserving the structural integrity of the latent representations in high-dimensional spaces. 
Other dimensionality reduction techniques, such as t-SNE or UMAP, while effective for visualization, are less suited for downstream optimization tasks due to their nonlinear transformations and stochastic nature.

Let \( \mathbf{W}_{\text{PCA}} \in \mathbb{R}^{d' \times d} \) denote the matrix of the top principal components, where \( d' \) is the reduced dimension and \( d \) is the original dimension. 
Each latent vector \( r_n \) is projected onto the lower-dimensional subspace as:
\begin{equation}
r'_n = \mathbf{W}_{\text{PCA}} r_n
\end{equation}
Similarly, the cluster centroids \( \mu_{r_n} \) and \( \mu_{c_k} \) are projected onto the same subspace as:
\begin{equation}
\mu'_{r_n} = \mathbf{W}_{\text{PCA}} \mu_{r_n}, \quad \mu'_{c_k} = \mathbf{W}_{\text{PCA}} \mu_{c_k}
\end{equation}
The number of retained principal components  \( dim\) is determined by ensuring a predefined threshold  \( T \)  of cumulative explained variance is met, as follows:
\begin{equation}
\label{eq:PCA_dims}
dim = \min \left\{ i \,\middle|\, \frac{\sum_{j=1}^{i} \frac{S_j^2}{m-1}}{\sum_{j=1}^{\text{max\_dim}} \frac{S_j^2}{m-1}} \geq T \right\} \quad \text{or} \quad dim = dim_{max} \text{ if no such } i \text{ exists}
\end{equation}
Here, \( S_j \) are the singular values from the Singular Value Decomposition (SVD) that is extracted from the PCA dimension reduction process, \( m \) is the number of samples, \( n \) is the number of available features, and \( \text{max\_dim} = \min(m, n) \) is the maximum number of possible components. 
\( T \) represents the threshold for cumulative explained variance, which we set to 0.95 based on initial investigations.

After applying PCA, we compute the \name{} loss using the reduced latent representations \( r'_n \) and \( \mu' \), including the distance information through a dense and efficient format.

\subsection{Individual Cluster Variance}
To ensure the model effectively adapts to the low dimensional representation for each cluster \( \sigma_C^2 \), the cluster variance is now computed based on the spread of points within each cluster individually. 
This allows the model to better account for the inherent variability and density differences within each cluster due to the filtered and compressed latent space information.
Specifically, the variance represents the average squared distance of the points \( r_i \) in cluster \( C \) from the cluster mean \( \mu_C \). 
The formula for calculating the variance for a cluster \( C \) is given by:
\begin{equation}
\label{eq:cluster_variance}
\sigma_C^2 = \frac{1}{|C| - 1} \sum_{r_i \in C} \| \mathbf{r}_i - \boldsymbol{\mu}_C \|^2
\end{equation}
Here, \( |C| \) denotes the number of points in cluster \( C \) and adjusts the degree of freedom.
\( r_i \) represents the position of the \( i \)-th point in the cluster and \( \mu_C \) is the centroid of the cluster from which we calculate the squared Euclidean distance between.
This dynamic variance allows the model to adapt to clusters with varying densities, meaning clusters that are more spread out will have larger variance and be more forcefully compressed compared to tighter clusters. 

\subsection{Dynamic Inter and Intra Cluster Balance}
Additionally, we introduce a hyperparameter $\beta$ in the denominator of the loss function.
The Magnet loss, as presented in \Cref{eq:magnet_base}, is composed of two main components: intra-cluster variance minimization, controlled by $\alpha$, and inter-cluster separation, now influenced by $\beta$. 
To balance these competing objectives, we developed dynamic strategies to adjust $\alpha$ and $\beta$ based on the current training epoch $E$, as formulated in \Cref{eq:hyper_schedule}. 
Initially, the focus is on achieving tight clustering of intra-class samples by assigning larger values to $\alpha$. 
Subsequently,  $\beta$  plays a greater role in encouraging separation between clusters.
The update rule for $\alpha$ follows an exponential decay schedule due to the simple subtraction as a margin term. 
It starts at a value of $1 +\alpha_0$, where $\alpha_0$ controls the initial strength, and gradually decreases by the factor $e^{-\frac{\text{E}}{1.05 \cdot \text{E}_{\text{total}}}}$ towards the commonly set value of 1.0 as training progresses. 
This approach ensures that the focus on minimizing intra-cluster variance gradually weakens as the model learns to form tighter clusters.

In contrast, the update rule for  $\beta$  follows a linear schedule, starting from $\beta_0$, and decreases linearly until it reaches zero after the first 20\% of the conventional training period. 
This linear decay ensures that the inter-cluster distance is encouraged early in training, but gradually becomes more significant as training progresses.
This schedule ensures that $\beta$ gradually diminishes and therefore increases the effect of distancing the clusters. 
This mechanism allows the model to be unconstrained by  $\beta$  in the early stages due to the multiplication factor and its initialization with 1.0, but ensures that the influence of the inter-cluster distance denominator increases for the overall loss.

Based on our experiments, the intra-cluster is mostly relevant during early epochs, whereas the exponential decrease of $\alpha$ helps to diminish the effect of intra-cluster variance distance later in the training progress.
However, the inter-cluster importance needs to be linearly increased by decreasing $\beta$ in order to move the clusters continuously and gradually farther apart till the model converges.

To ensure numerical stability across our experiments, we added $\epsilon$ and set it to $1 \cdot 10^{-8}$ as a diminutive constant to prevent division by zero and underflow of floating point numbers.
Furthermore, $\epsilon$ is introduced as a minimum value for $\beta$  to prevent it from diminishing completely and to avoid destabilization of the training process.
\begin{equation}
\label{eq:hyper_schedule}
\begin{aligned}
    \alpha &= 1 + \alpha_0 \cdot e^{-\frac{\text{E}}{1.05 \cdot\text{E}_{\text{total}}}} \quad & 
    \beta &= \max\left(\beta_0 \cdot \left(1 - \frac{\text{E}}{0.2 \cdot \text{E}{\text{total}}}\right), \epsilon\right)
\end{aligned}
\end{equation}
\section{Full Potential of Latent Boost}
Combining the previously introduced adaptations to the superior Magnet loss from our experiments, we can form the \name{} loss as stated in \Cref{eq:latent_boost}.
Finally, the \name{} loss is balanced through the $\lambda$ hyperparameter with the cross-entropy loss to enhance multi-class classification.
The full equation as part of the main contribution of this work is expressed as follows:
\begin{equation}
\label{eq:latent_boost}
\begin{gathered}
\mathcal{L}_{\text{LB}} = \frac{1}{N} \sum_{n=1}^{N} \left\{ -\log \left( \frac{e^{- \frac{1}{2 \sigma_{C_k}^2} \| \mathbf{r}'_n - \boldsymbol{\mu}'_{r_n} \|^2_2 - \alpha}}{\sum_{c \neq C(r_n)} \sum_{k=1}^{K} e^{-\frac{1}{2 \sigma_{C_k}^2} \| \mathbf{r}'_n - \boldsymbol{\mu}'_{c_k} \|^2_2 \cdot \beta }} + \epsilon \right) \right\}, \quad
\mathcal{L}_{\text{cross-entropy}} = -\sum_{\forall x} p(x) \log(q(x))
\\
\\
 \mathcal{L}_{total} = \lambda \cdot \mathcal{L}_{\text{LB}} + (1 - \lambda) \cdot \mathcal{L}_{\text{cross-entropy}} 
\end{gathered}
\end{equation}

\begin{figure}[ht] 
    \centering
    \includegraphics[width=\textwidth]{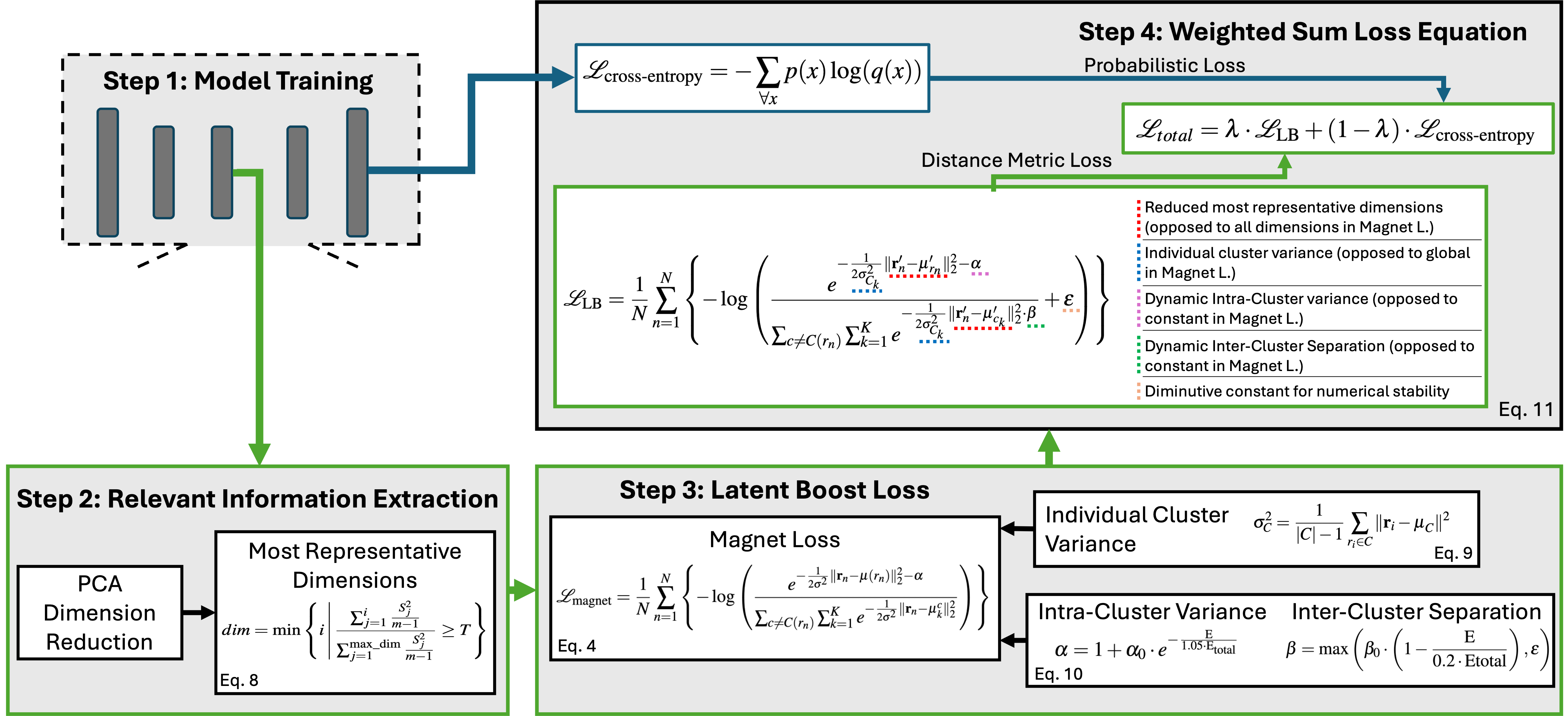} 
    \caption{Flowchart of the \name{} approach, summarizing the mathematical steps to embed distance-metric information into the classic probabilistic training.}
    \label{fig:math_flowchart}
    \vspace{-1em}
\end{figure}

In \Cref{fig:math_flowchart}, the whole process of \name{}, applied within each batch iteration, is visualized comprehensively.
Within the traditional model training, we take the latent space data from intermediate layers. 
We reduce the dimensions using PCA and extract the most representative ones based on \Cref{eq:PCA_dims}.
Afterwards, we calculate the \name{} based on the Magnet loss from \Cref{eq:magnet_base} and the modifications of \Cref{eq:cluster_variance} and \Cref{eq:hyper_schedule}.
In the final step, we combine the classic Cross-Entropy loss with the \name{} part through the weighted sum loss from \Cref{eq:latent_boost} with the $\lambda$ parameter.

\subsection{Improved Classification Performance}

To evaluate our approach, we conducted the same experiment, based on our three datasets and model combinations, with the adapted \name{} distance metric.
The selected $\lambda$ values mimic the range of the preliminary experiment from \Cref{prelim_evaluation} starting from 0.1 to 0.9.
The results can be found in \Cref{tab:eval_LB_full} and show the performance of \name{} across three datasets and model combinations. 
For Fashion MNIST, the best performance is achieved at $\lambda = 0.75$ with an accuracy of 90.86\% and a Micro-F1 score of 0.9038. 
Similarly, on CIFAR-10, $\lambda = 0.75$ gives the highest accuracy (88.44\%) and Micro-F1 score (0.8843), while higher values of $\lambda$ slightly reduce performance. 
On CIFAR-100, the performance is less sensitive to $\lambda$, with the best results at $\lambda = 0.5$. 
Overall, $\lambda$ values between 0.5 and 0.75 consistently deliver the best results, indicating an optimal balance for beneficial performance while structuring the latent spaces.
\begin{table}[ht]
\footnotesize
    \centering
    \begin{tabular}{c|cc|cc|cc}
    \hline
        \multirow{2}{*}{\textbf{$\boldsymbol{\lambda}$}} & \multicolumn{2}{c|}{\textbf{Fashion MNIST  (Mean ± Std)}} & \multicolumn{2}{c|}{\textbf{CIFAR-10 (Mean ± Std)}} & \multicolumn{2}{c}{\textbf{CIFAR-100 (Mean ± Std)}} \\
        \cline{2-7}
        & \textbf{Accuracy} & \textbf{Micro-F1} & \textbf{Accuracy} & \textbf{Micro-F1} & \textbf{Accuracy} & \textbf{Micro-F1} \\
        \hline
        0.1  & 88.47 ± 0.16  & 0.8892 ± 0.0019  & 83.11 ± 0.19  & 0.8342 ± 0.0018  & 60.34 ± 0.19  & 0.6112 ± 0.0015 \\
        0.2  & 89.03 ± 0.12  & 0.8905 ± 0.0017  & 84.29 ± 0.16  & 0.8431 ± 0.0017  & 61.45 ± 0.12  & 0.6170 ± 0.0016 \\
        0.25 & 89.36 ± 0.14  & 0.8923 ± 0.0020  & 85.22 ± 0.20  & 0.8464 ± 0.0019  & 62.21 ± 0.17  & 0.6215 ± 0.0018 \\
        0.3  & 89.62 ± 0.09  & 0.8927 ± 0.0015  & 85.64 ± 0.15  & 0.8485 ± 0.0015  & 62.67 ± 0.22  & 0.6258 ± 0.0015 \\
        0.4  & 90.05 ± 0.12  & 0.8983 ± 0.0018  & 86.05 ± 0.13  & 0.8510 ± 0.0016  & 63.02 ± 0.33  & 0.6280 ± 0.0017 \\
        0.5  & 90.47 ± 0.19  & 0.9005 ± 0.0019  & 86.85 ± 0.12  & 0.8554 ± 0.0015  & \textbf{63.04 ± 0.48}  & \textbf{0.6288 ± 0.0018} \\
        0.6  & 90.15 ± 0.13  & 0.9021 ± 0.0022  & 87.43 ± 0.11  & 0.8598 ± 0.0013  & 62.83 ± 0.31  & 0.6275 ± 0.0016 \\
        0.7  & 90.09 ± 0.21  & 0.9015 ± 0.0018  & 88.15 ± 0.24  & 0.8641 ± 0.0014  & 62.45 ± 0.27  & 0.6268 ± 0.0015 \\
        0.75 & \textbf{90.86 ± 0.21} & \textbf{0.9038 ± 0.0025} & \textbf{88.44 ± 0.28} & \textbf{0.8843 ± 0.0024} & 62.30 ± 0.14  & 0.6255 ± 0.0017 \\
        0.8  & 89.76 ± 0.14  & 0.8952 ± 0.0016  & 87.92 ± 0.22  & 0.8803 ± 0.0015  & 62.00 ± 0.18  & 0.6232 ± 0.0018 \\
        0.9  & 88.53 ± 0.15  & 0.8902 ± 0.0018  & 86.30 ± 0.18  & 0.8725 ± 0.0018  & 60.78 ± 0.11  & 0.6204 ± 0.0016 \\
        \hline
    \end{tabular}
    \caption{Complete performance evaluation of \name{} across the full set of $\lambda$ values.}
    \label{tab:eval_LB_full}
    \vspace{-1em}
\end{table}

Additionally, as shown in \Cref{tab:lb_performance}, we track the accuracy, Micro-F1 Score, and the duration of training epochs.
We calculated the average and standard deviation across our five experiment trials to compare our baseline without distance metrics and the classic Magnet loss addition. 
For the \name{}, we added two versions, first without the addition of PCA to isolate the effects of variance and the dynamic update rule and finally the full \name{} with all features.
With that, we can manifest the performance of \name{} applied to the full high-dimensional space and obtain insights into the effects of extracting the necessary information from the dimension reduction process.

\name{} proves to consistently outperform the baseline and the classic Magnet loss results from the previous experiments of \Cref{tab:baseline}.
For Fashion-MNIST and CIFAR-10, $\lambda$ selection of 0.75 and 0.5 for the CIFAR-100 obtained the best performance equal to the original Magnet loss results.
The Fashion MNIST shows a 2.56\% increase in accuracy and a 1.93\% increase in F1 Score, while CIFAR-10 and CIFAR-100 exhibit comparable improvements of around 2-3\%. 
These gains are accompanied by reduced standard deviations, indicating a more stable and reliable convergence due to the additional structural information from latent representation. 
The tighter variability in performance indicates that \name{} is more consistent across datasets. 

\begin{table}[ht]

    \centering
    \footnotesize
    
    \begin{tabular}{c|c|c|c|c|c|c}
    \hline
        \textbf{Dataset} & \textbf{Metric} & \textbf{Baseline} & \textbf{Magnet} & \textbf{Latent Boost w/o PCA} & \textbf{Latent Boost} & \textbf{Improvement (\%)} \\ \hline
        \multirow{3}{*}{\textbf{Fashion MNIST}} & Accuracy & 88.59 ± 0.15 & 89.52 ± 0.34 & 88.12 ± 0.29 & \textbf{90.86 ± 0.21} & 2.56\%  \\ 
                                                & Micro-F1 & 0.8867 ± 0.002 & 0.8946 ± 0.004 & 
                                                0.8971 ± 0.004 & \textbf{0.9038 ± 0.005} & 1.93\% \\
                                                & Nr. Epochs & 40.67 ± 4.19 & 37.67 ± 6.18 & 
                                                34.86 ± 4.26    & \textbf{34.67 ± 3.09} & -14.76\% \\ \hline
        \multirow{3}{*}{\textbf{CIFAR-10}}      & Accuracy & 85.88 ± 0.40 & 87.36 ± 0.82 & 88.12 ± 0.47 & \textbf{88.44 ± 0.28} & 2.98\% \\ 
                                                & Micro-F1 & 0.8586 ± 0.0042 & 0.8738 ± 0.0081 &  0.8811 ± 0.0040 & \textbf{0.8843 ± 0.0024} & 2.99\% \\ 
                                                & Nr. Epochs & 76.33 ± 8.65 & 72.67 ± 1.70 & 71.94 ± 3.72 & \textbf{66.33 ± 4.50} & -13.09\% \\ \hline
        \multirow{3}{*}{\textbf{CIFAR-100}}     & Accuracy & 61.77 ± 0.49 & 62.75 ± 0.51 & 63.05  ± 0.52& \textbf{63.04 ± 0.48} & 2.06\% \\ 
                                                & Micro-F1 & 0.6163 ± 0.0064 & 0.6256 ± 0.0057 & 0.6218 ± 0.0059 & \textbf{0.6288 ± 0.0051} & 2.03\% \\ 
                                                & Nr. Epochs & 86.67 ± 9.46 & 69.67 ± 8.73 & 69.23 ± 8.92 & \textbf{68.67 ± 6.02} & -20.74\% \\
                                                \hline
    \end{tabular}
    \caption{Accuracy ($\uparrow$), Micro-F1 Score ($\uparrow$), and epoch duration ($\downarrow$) between baseline ($\lambda=0$), standard Magnet loss, the \name{} without PCA and the final \name{} with all features on the unseen test dataset (best $\lambda$ selection); percentage improvement compares \name{} with the baseline for each metric.}
    \label{tab:lb_performance}
    \vspace{-1em}
\end{table}

Another advantage of \name{} is its faster and more stable convergence next to the model performance. 
The number of epochs required for Fashion MNIST training is reduced by around 14\%, while CIFAR-10 and CIFAR-100 show reductions of around 13\% and 21\%, respectively. 
This reduction not only speeds up training but also offers potential sustainability benefits by decreasing computational resources and energy consumption. 

As stated before, the threshold for selecting the number of principal components was decided to be at 95\% of cumulative explained variance.
For Fashion-MNIST, the number of principal components ranged around 45 to 50, with slight degradation to 40 components when the latent representation reaches a sufficient structure.
For CIFAR-10, the number of principal components started around 65 and decreased to 40 components, stating that the latent representation is well structured throughout the epochs.
For the final CIFAR-100 experiment, the principal components used were around 100 to 110 without a decent decrease over time.
That being said, it reflects our findings of struggling to structure the large latent representation properly.

\subsection{Improved Latent Representation Interpretability}
We define interpretable classification as one that achieves not only the correct assignment of discrete data points to their respective classes but also induces a structured organization of latent representations, where clusters corresponding to each class are maximally separated and distinct.

\begin{figure}[!ht]
    \centering
       
    \begin{subfigure}{0.32\textwidth}
        \centering
        \includegraphics[width=0.9\textwidth]{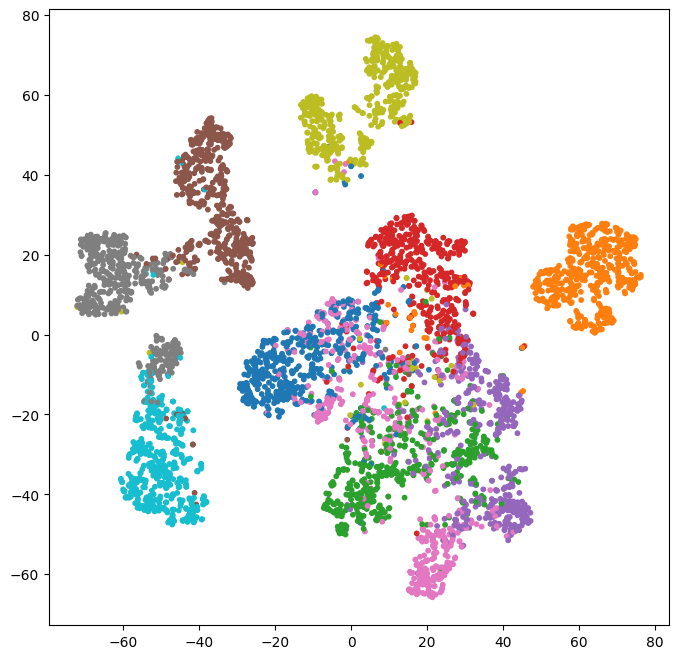}
        \caption{Fashion MNIST Baseline}
    \end{subfigure}
    \hfill
    \begin{subfigure}{0.32\textwidth}
        \centering
        \includegraphics[width=0.9\textwidth]{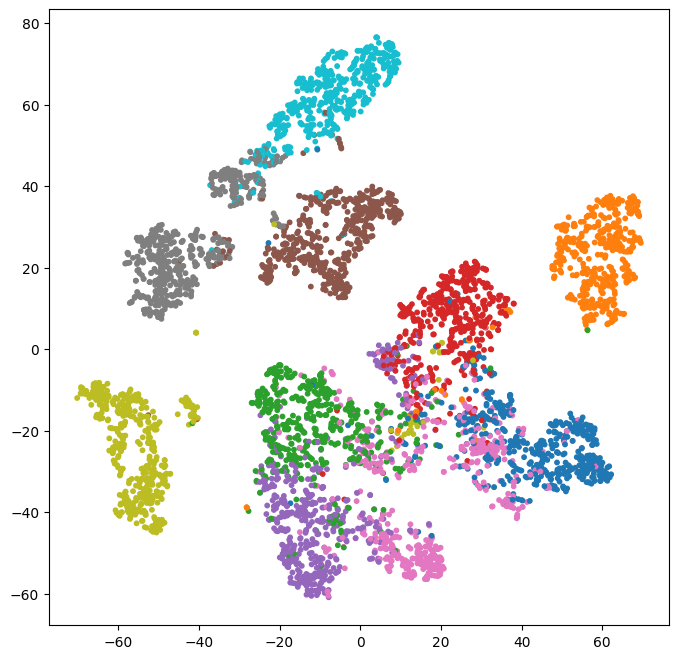}
        \caption{Fashion MNIST Magnet}
    \end{subfigure}
    \hfill
    \begin{subfigure}{0.32\textwidth}
        \centering
        \includegraphics[width=0.9\textwidth]{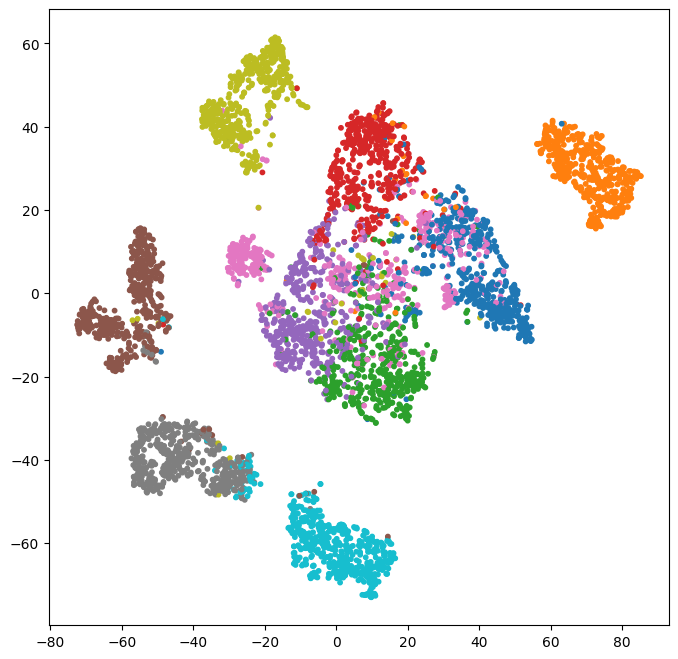}
        \caption{Fashion MNIST \name{}}
    \end{subfigure}
    

    \begin{subfigure}{0.32\textwidth}
        \centering
        \includegraphics[width=0.9\textwidth]{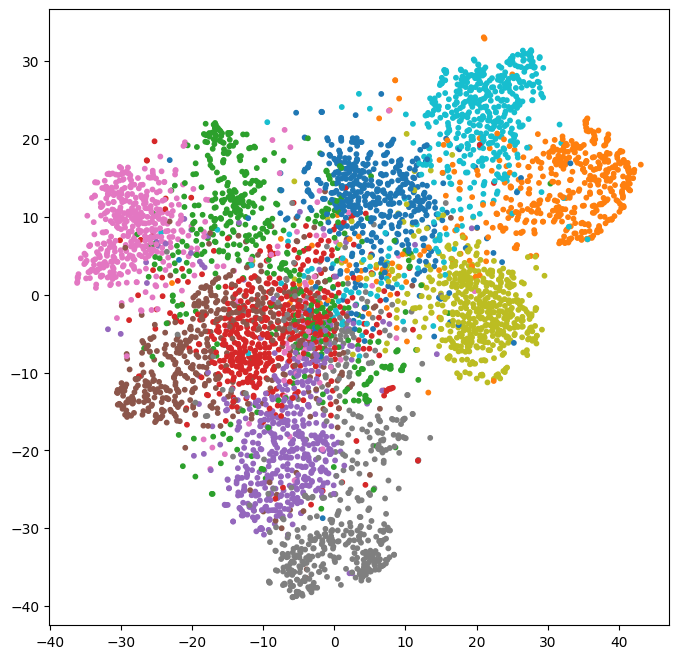}
        \caption{CIFAR-10 Baseline}
    \end{subfigure}
    \hfill
    \begin{subfigure}{0.32\textwidth}
        \centering
        \includegraphics[width=0.9\textwidth]{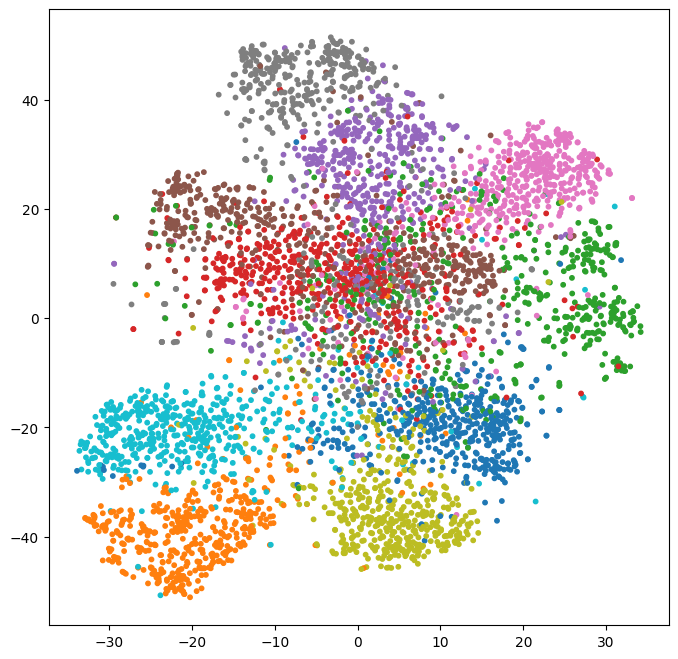}
        \caption{CIFAR-10 Magnet}
    \end{subfigure}
    \hfill
    \begin{subfigure}{0.32\textwidth}
        \centering
        \includegraphics[width=0.9\textwidth]{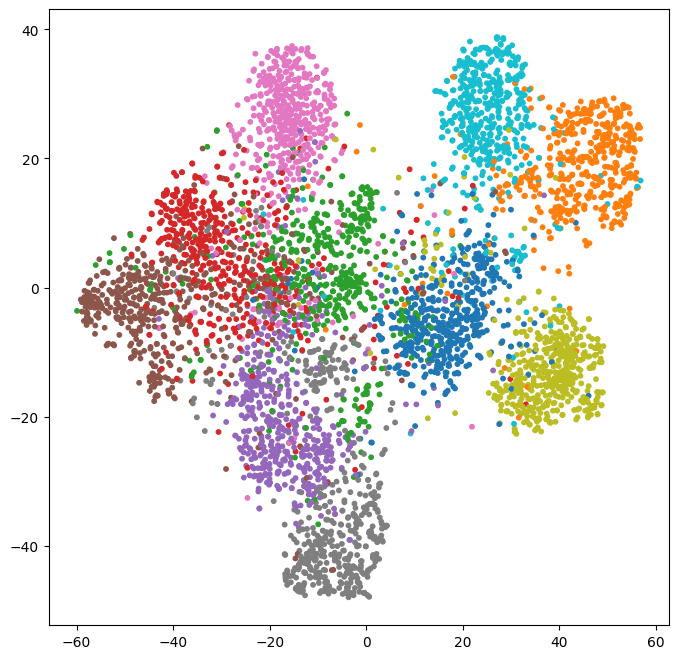}
        \caption{CIFAR-10 \name{}}
    \end{subfigure}
    
    
    \begin{subfigure}{0.32\textwidth}
        \centering
        \includegraphics[width=0.9\textwidth]{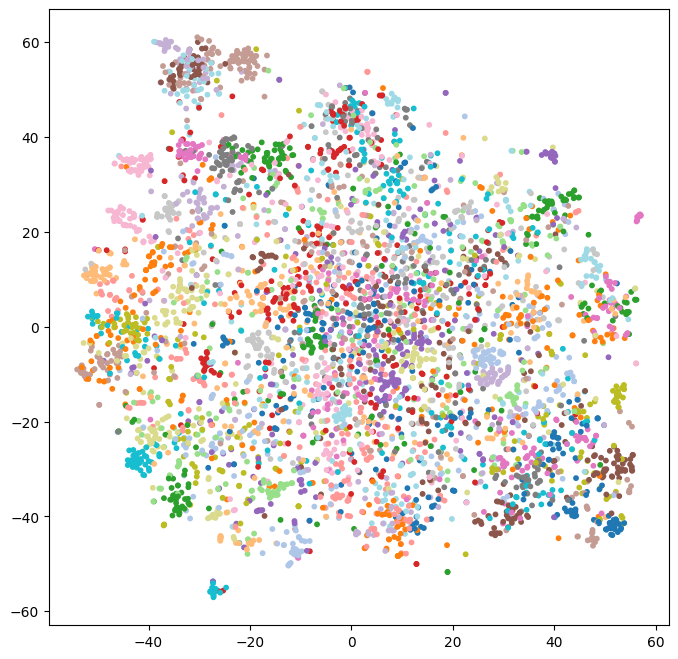}
        \caption{CIFAR-100 Baseline}
    \end{subfigure}
    \hfill
    \begin{subfigure}{0.32\textwidth}
        \centering
        \includegraphics[width=0.9\textwidth]{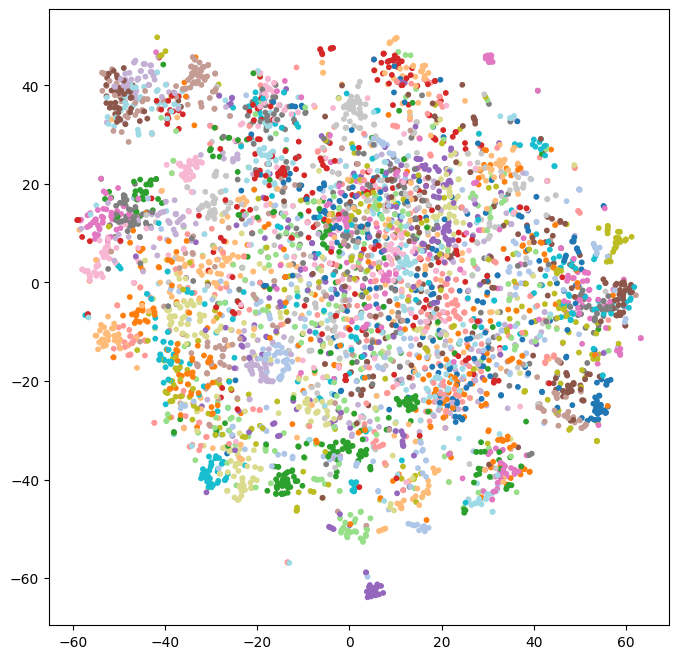}
        \caption{CIFAR-100 Magnet}
    \end{subfigure}
    \hfill
    \begin{subfigure}{0.32\textwidth}
        \centering
        \includegraphics[width=0.9\textwidth]{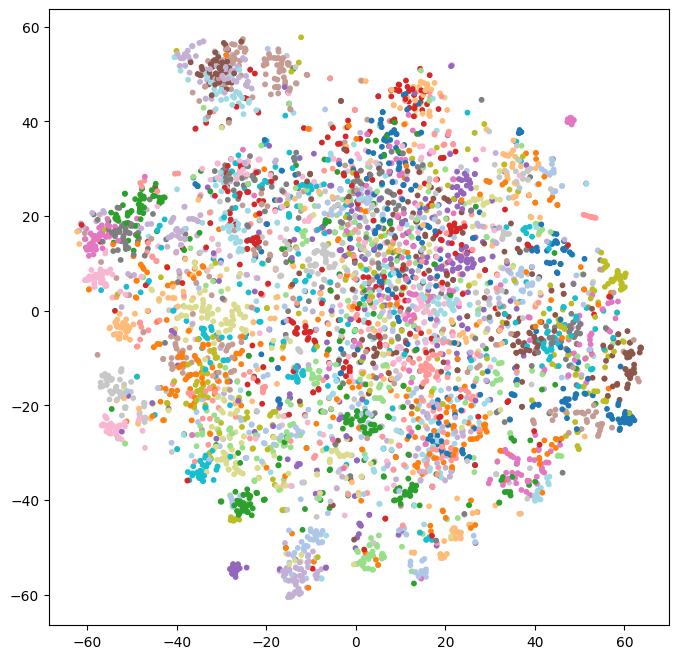}
        \caption{CIFAR-100 \name{}}
    \end{subfigure}
    \caption{Comparison of baseline ($\lambda$=0), standard Magnet loss, and our \name{} approach across the three experiment setups.}
    \label{fig:comparison_latent}
    \vspace{-1em}
\end{figure}

\textbf{Qualitative Inspection: }
In \Cref{fig:comparison_latent}, we plotted the 2-dimensional representation of the high-dimensional latent space for each of our experiments.
We passed the test dataset of each experiment through the model and extracted the latent representation the same way as previously proposed.
However, we do not calculate the \name{} metric but rather utilize the data and shrink its dimension with classic and state-of-the-art TSNE \cite{van2008visualizing} for visualization purposes.
We utilized the TSNE according to best practices, keeping hyperparameters in its standard initialization and fixing the random seed.
The visualizations from left to right show the latent representation for the best trial of baseline training without any distance metric information, the Magnet loss training from our preliminary experiments, and the \name{} approach. 
From top to bottom, we show each of the experiment models and dataset combinations.

Starting with the Fashion MNIST, even though the clusters were separated properly in the baseline already, we can see that in the \name{} clusters are formed with more concentrated density and classes such as \textit{brown} and \textit{grey} are well separated.
In the scenario with CIFAR-10, a similar trend can be recognized.
Even though there is much more confusion within the latent representation compared to the Fashion MNIST, clusters are more dense in the final stage of \name{}.
Additionally, the main confusion between \textit{green} and \textit{red} class from the baseline could partially be resolved.
Some classes are much better formed for clustering and boundaries between the classes can be recognized stronger.
For the final experiment on CIFAR-100, the visualization is not meaningful enough to show a clear separation of clusters.
This however may be the issue of visualizing the 100 classes with different color shades.
Only the surrounding areas of the latent representation form small clusters without significant visible impact.

\textbf{Quantitative Measurement:}
Since TSNE compresses the latent space and therefore loses the detailed information from the full dimensional representation, the visualizations can only be utilized for visual cross-comparison but do not suffice for compelling quantitative evaluation.
To quantify the density of clusters and their separation from each other in the original dimension, we selected the Silhouette Score to measure the quality of the latent representation.
Originally proposed by \cite{rousseeuw1987silhouettes}, the Silhouette Score is calculated for each data point by comparing the \textbf{cohesion} within its own cluster and the \textbf{separation} from the nearest neighboring cluster. 
It has been widely used for quantifying representation interpretability in recent works \cite{bagirov2023finding,januzaj2023determining, du2024multi}.
The score ranges from \(-1\) to \(1\), with -1 indicating the location or assignment of samples to the wrong cluster, whereas 0 is between clusters and 1 is the best-case with clear allocation to the correct cluster.
For a given data point \(i\) in a cluster \(C_i\), the Silhouette Score \(s(i)\) is calculated following \Cref{eq:silhouette}.
$a(i)$ represents the cohesion as to how closely related a data point is to its own cluster, whereas $b(i)$ represents the separation, meaning the distance between a data point to its nearest neighbor cluster.
To evaluate the impact of different training approaches on the latent representation, we calculated the Silhouette Scores for the previously discussed latent representations in \Cref{fig:comparison_latent}. 
The results, shown in \Cref{table:silhouette}, highlight that both the Magnet loss and \name{} methods improve cluster separation on the full dimension, relative to the baseline.
\name{} yields the highest score in all cases, with particularly strong improvements observed for CIFAR-10.
\begin{equation}
\label{eq:silhouette}
\centering
\begin{gathered}
    \text{Silhouette Score} = \frac{1}{N} \sum_{i=1}^{N} \frac{b(i) - a(i)}{\max(a(i), b(i))} \\
    a(i) = \frac{1}{|C_i| - 1} \sum_{j \in C_i, j \neq i} d(i, j) \quad \quad 
    b(i) = \min_{C \neq C_i} \left( \frac{1}{|C|} \sum_{j \in C} d(i, j) \right) 
\end{gathered}
\end{equation}
The results reflect the preliminary experiment hypothesis across the three datasets.
Checking the improvement between baseline and \name{}, the Fashion MNIST experiment showed slight improvement, since the baseline latent space was already well separated.
The CIFAR-10 achieved the greatest improvement in the dimensional representation with around 168 \%, restating the model's complexity potential for improvement, especially in the high dimensions.
For the CIFAR-100, even though some minor improvement could be recognized, the \name{} approach does not sufficiently support the training as expected.
Since the Silhouette Score ranges in negative values for CIFAR-100, major confusion between the large set of classes still dominates the classification despite the slight increase.

\begin{table}[ht]
    \centering
    \footnotesize
   
    \begin{tabular}{c|c|c|c|c}
    \hline
        \textbf{Dataset} & \textbf{Baseline} & \textbf{Magnet} & \textbf{Latent Boost} & \textbf{Improvement} (\%) \\ \hline
        \textbf{Fashion MNIST} & 0.347 & 0.375 & \textbf{0.395} & 13.84\% \\ 
        \textbf{CIFAR-10}      & 0.131 & 0.186 & \textbf{0.351} & 167.94\% \\ 
        \textbf{CIFAR-100}     & -0.280 &  -0.264 & \textbf{-0.250} & 10.71\% \\ 
        \hline
    \end{tabular}
     \caption{Silhouette Score to quantify cluster separation (larger values in range -1 to 1 represent greater separation); calculated on the models' latent representation without compression from the unseen test dataset; the percentage improvement compares \name{} with the baseline.}
    \label{table:silhouette}
    \vspace{-1em}
\end{table}

While the Silhouette Score captures the quality of latent representation, it is important to note that the model is not directly trained on the latent representation data. 
Instead, the calculated metrics are injected as additional information through the loss function, helping the model to indirectly improve its internal representation structures. 
This approach leverages the power of the latent representation without requiring explicit supervision, allowing the model to develop more meaningful features.

\section{Discussion}
\subsection{Overfitting Risk and Mitigation}
Mitigating overfitting is essential to ensure that the benefits of \name{} extend beyond the training dataset. 
During the evaluation, \name{} computations were performed exclusively on the test dataset in a batch-wise manner to maintain a strict separation between the training and evaluation phases, avoiding any potential information leakage. 
Early stopping was employed as another measure to prevent overfitting, halting training when validation performance plateaued. 
This not only curtailed unnecessary epochs but also reduced computational overhead.
Dropout regularization was also applied to deactivate neurons stochastically during training, introducing variability and encouraging the model to distribute learning across multiple pathways.
Together, these strategies ensured that \name{} effectively balanced high training accuracy with robust generalization, making it suitable for deployment in real-world scenarios.
Although out-of-distribution testing was not explicitly conducted, it could serve as a valuable future regularization strategy to assess whether the model generalizes effectively or overfits to the training data. 
Out-of-distribution testing evaluates model performance on data that deviates from the training distribution, offering insights into the model’s latent space organization and robustness under varied conditions.

\subsection{Dynamic Lambda Variations }
The weighting factor $\lambda$ plays a critical role in balancing the focus on classification performance and interpretability in order to maximize them both in a symbiotic way.
We empirically selected the $\lambda$ values in the range 0.5 to 0.75 commonly gained the greatest enhancement.
However, we kept $\lambda$ values constant across experiments.
Rudimentary trials with dynamic adjustment of $\lambda$ showed instability and computational challenges despite the potential for deeper fine-tuning.
While the constant approach simplifies implementation, it also limits the adaptability of the method. 
Future enhancements might involve more automated strategies to dynamically adjust $\lambda$ without compromising training stability, potentially leveraging reinforcement learning or other adaptive optimization techniques.

\subsection{Training Resource Efficiency Benefits}
\name{} requires several normal epochs with only the probabilistic cross-entropy loss to form initial clusters for further effective manipulation and refinement of the cluster structures in the latent space. 
While \name{} in general introduces additional computational overhead per epoch due to its latent space calculations, this is offset by faster convergence rates that reduce overall training time, especially due to the dimension reduction to thin out the process.
The utilized PCA for dimension reduction is computationally efficient and significantly reduces the size of the latent representations, making downstream computations faster and less resource-intensive.
\name{} only takes the number of dimensions determined by \cref{eq:PCA_dims}, whereas in the SOTA, Magnet loss computes the loss term with all upstream feature dimensions.
Notwithstanding, its benefits are reduced in scenarios where training is inherently limited to a small number of epochs. 

\subsection{Cluster Separability}
Our method assumes that class clusters form hyper-spherical structures within the multi-dimensional latent space. 
This assumption may not hold for all datasets, particularly those with irregular, highly overlapping or hierarchical cluster geometries as recognized in CIFAR-100. 
Exploring alternative clustering strategies or loss formulations tailored to diverse latent space distributions could further enhance the method’s performance and applicability. 
\name{} also assumes that classes are generally independent in the feature space, allowing clusters to be separable. 
However, in scenarios where classes exhibit cross-dependencies in certain feature dimensions, disentangling latent representations becomes more challenging. 
This limitation may explain the relatively modest gains observed for CIFAR-100. 
Future research could explore clustering techniques that account for hierarchical relationships or cross-class dependencies to improve performance further.


\section{Conclusion}

Overall, \name{} explicitly embeds the classification objective into the latent layer of an otherwise black-box model in the form of an additional loss term. 
This integration encourages the formation of better-separated clusters in the target latent space while simultaneously optimizing for prediction performance. 
As a result, the method achieves drastically improved latent interpretability, enhanced classification outcomes, and more efficient training convergence.
While the uplift in classification performance is moderate, it is consistent across all benchmarks, exhibiting minimal deviation. 
Additionally, our approach facilitates faster convergence, aligning with the goals of sustainable machine learning by reducing both training time and computational resource consumption.

In conclusion, \name{} represents a significant advancement in harmonizing interpretability, efficiency, and performance within supervised learning. 
Its ability to effectively structure latent representations underscores its potential as a versatile solution to the interpretability challenges posed by modern machine-learning black-box models.

\bibliography{references}

\section*{Acknowledgements}
This work is supported by the European Union’s Horizon Europe research and innovation program (HORIZON-CL4-2021-HUMAN-01) through the ”SustainML” project (grant agreement No. 101070408).
The Carl-Zeiss Stiftung also funded it under the Sustainable Embedded AI project (P2021-02-009).

\section*{Author contributions statement}

D.G., B.Z.: Conceptualization, Methodology, Experiment and Writing the Original Draft;
M.L.: Experiment;
P.L.: Funding Acquisition and Supervision;


\subsection*{Availability of Data and Materials}
The datasets and model architectures analyzed during this work are publicly available for reproduction of this approach.

\section*{Additional information}
\textbf{Competing interests:} The authors declare no competing financial and non-financial interests.

\end{document}